\documentclass[11pt]{article}
\usepackage[margin=1in]{geometry}
\usepackage{amsmath,amssymb,booktabs,multirow,graphicx,hyperref,microtype,array}
\usepackage{enumitem}
\usepackage{caption}
\usepackage{authblk}
\hypersetup{colorlinks=true,linkcolor=blue,citecolor=blue,urlcolor=blue}
\title{CacheDyG: Decoupling Temporal Propagation for Efficient Dynamic Graph Learning}
\author[1]{PinHeng Zong}
\author[1]{Ye Yuan\thanks{Corresponding author: yuanyekl@swu.edu.cn}}
\affil[1]{College of Computer and Information Science, Southwest University, Chongqing, China\\\texttt{zongph@email.swu.edu.cn, yuanyekl@swu.edu.cn}}
\date{}

\begin{document}
\maketitle
\begin{center}
\small\textit{Preprint. Accepted for presentation/publication at ADMA 2026. This manuscript is not the Springer Version of Record.}
\end{center}

\begin{abstract}
Dynamic graphs are widely used to model time-evolving relational systems in real-world applications. Dynamic graph neural networks provide an effective framework for capturing both structural dependencies and temporal dynamics in such data. However, they typically intertwine temporal graph propagation with every optimization epoch and often maintain large trainable representations for each node-time pair. This design repeatedly recomputes largely unchanged historical structures, leading to substantial training and parameter overhead. To address this critical issue, we propose CacheDyG, a Cache-refine framework for efficient Dynamic Graph learning. Specifically, it decouples temporal propagation from routine parameter updates by constructing a time-ordered temporal dependency cache that stores graph-aware node-time representations in non-trainable buffers. During standard training epochs, CacheDyG reads from the cache and updates only a lightweight cache refiner, an adaptive residual gate, and the link predictor. Selective cache refresh further keeps cached representations aligned with the supervised objective while avoiding epoch-wise sparse propagation. Experiments on five dynamic graph benchmarks show that CacheDyG adopts substantially fewer trainable parameters and lower runtime to obtain more competitive predictive performance than baselines. These results demonstrate that cache-based decoupling provides an effective principle for scalable dynamic graph learning.
\end{abstract}

\noindent\textbf{Keywords:} Dynamic graph learning; Dynamic graph neural networks; Link prediction; Cache mechanism; Decoupled parameter update

\section{Introduction}
Dynamic graphs provide a natural representation for relational systems whose interactions evolve over time, including social networks, recommender systems, transaction networks, communication systems, and biological interaction networks~\cite{kazemi2020survey,yuan2020generalized}. Unlike static graphs, dynamic graphs exhibit both structural dependencies among entities and temporal dependencies across different stages of evolution. Learning from such data therefore requires models to preserve graph-structured information while capturing how relational patterns change over time. As real-world dynamic graphs grow in both scale and temporal duration, computationally and memory-efficient learning becomes increasingly important.

Dynamic graph neural networks provide an effective framework for jointly modeling structural and temporal dependencies. Snapshot-based methods, such as DySAT~\cite{sankar2020dysat}, EvolveGCN~\cite{pareja2020evolvegcn}, and VGRNN~\cite{hajiramezanali2019vgrnn}, combine graph propagation with temporal attention, recurrent parameter evolution, or latent-state transitions. Continuous-time methods, including DyRep~\cite{trivedi2019dyrep}, JODIE~\cite{kumar2019jodie}, TGAT~\cite{xu2020tgat}, TGN~\cite{rossi2020tgn}, CAW~\cite{wang2021caw}, GraphMixer~\cite{cong2023graphmixer}, and DyGFormer~\cite{yu2023dygformer}, learn from timestamped interaction streams through event-driven memory, temporal neighborhoods, or sequence encoders. Although these methods have achieved strong predictive performance, they are generally designed to enhance representation expressiveness rather than to eliminate computational redundancy during optimization.

This redundancy is particularly pronounced in transductive snapshot-based training. The historical graph sequence remains largely unchanged across optimization epochs, yet temporal graph propagation is often recomputed together with every gradient update. Moreover, some models maintain directly trainable representations for individual node-time pairs, causing the parameter size to grow with both the number of nodes and the number of snapshots. These two design choices create distinct but related efficiency bottlenecks: repeated sparse propagation over reusable historical structures increases runtime, while large node-time parameter tables increase memory consumption and optimization cost. As dynamic graphs become larger or span longer time horizons, such tightly coupled training pipelines can become unnecessarily expensive or even infeasible.

Our key observation is that temporal graph propagation and supervised parameter optimization do not need to operate at the same frequency. Historical graph structures can be propagated occasionally and stored as reusable representations, whereas lightweight task-specific modules can still be updated at every epoch. Based on this observation, we propose CacheDyG, a cache-refine framework that decouples temporal propagation from routine parameter updates. CacheDyG first constructs a time-ordered Temporal Dependency Cache using temporal mixing, so that each snapshot aggregates only current and past information. Graph-aware propagation is then performed to produce reusable node-time representations stored in non-trainable buffers. During ordinary training epochs, the model reads these cached representations and updates only a compact node-domain frequency-domain cache refiner, an adaptive residual gate, and a lightweight link predictor. A selective cache-refresh mechanism periodically incorporates task-aware refined representations back into the cache, keeping it aligned with the supervised objective without returning sparse graph propagation to the inner optimization loop.

This cache-based separation changes the role of graph propagation in dynamic graph learning. Instead of being repeatedly executed as part of every parameter update, propagation becomes an amortized representation-construction step, while optimization focuses on correcting and scoring the cached signal. The resulting architecture reduces trainable parameter growth, avoids repeated computation over unchanged historical snapshots, and preserves the structural and temporal evidence required for future-link prediction.

We evaluate CacheDyG on the task of one-step future-link prediction across five dynamic graph benchmarks, considering average precision, ROC-AUC, trainable parameter count, and runtime. The results show that CacheDyG achieves the best average precision on all evaluated datasets while using only 19.106K trainable parameters. It also obtains the lowest runtime and remains feasible on larger datasets where memory-intensive baselines fail. Ablation and sensitivity studies further demonstrate that temporal mixing, cache-style propagation, lightweight refinement, and selective refresh jointly contribute to the final accuracy-efficiency trade-off.

\paragraph{Contributions.} The main contributions of this work are summarized as follows:
\begin{enumerate}[leftmargin=*]
\item We identify a fundamental efficiency mismatch in transductive snapshot-based dynamic graph learning: reusable historical graph structures are repeatedly propagated during optimization, while directly trainable node-time representations can cause parameter and memory costs to grow rapidly with graph size and temporal length.
\item We propose CacheDyG, a cache-refine framework that decouples temporal graph propagation from routine parameter updates through a Temporal Dependency Cache, non-trainable graph-aware node-time buffers, lightweight frequency-domain refinement, adaptive residual gating, and selective cache refresh.
\item We provide computational and empirical evidence that the proposed decoupling amortizes sparse temporal propagation and yields a favorable accuracy-efficiency trade-off on five dynamic graph datasets in terms of predictive performance, trainable parameter size, runtime, and scalability.
\end{enumerate}

\section{Related Work}
\subsection{Dynamic Graph Learning}
Dynamic graph learning studies how to represent and predict relations in networks whose edges, attributes, or node states evolve over time~\cite{kazemi2020survey,wang2026gtcn,yuan2026highorder,yuan2025nodecollab,wang2025gta2t}. Earlier dynamic embedding methods learn temporally varying node representations from snapshot sequences or interaction histories~\cite{yuan2023kalman,yuan2020ecai,yuan2026qos}. DynamicTriad~\cite{zhou2018dynamictriad} models triadic closure processes, DynGEM~\cite{goyal2018dyngem} incrementally updates autoencoder-based graph embeddings, dyngraph2vec~\cite{goyal2020dyngraph2vec} captures temporal graph evolution with deep representation learning, and CTDNE~\cite{nguyen2018ctdne} extends random-walk-based embeddings to continuous-time dynamic networks. These methods establish the importance of temporal dependency for downstream tasks such as link prediction, but they generally focus on learning temporal embeddings rather than explicitly reducing repeated graph propagation in snapshot-based training.

Recent dynamic graph learning also includes interaction-sequence models that encode temporal events or historical neighborhoods for future-edge prediction. JODIE~\cite{kumar2019jodie} learns trajectories of user-item embeddings, DyRep~\cite{trivedi2019dyrep} models evolving interaction processes, CAW~\cite{wang2021caw} represents temporal networks through causal anonymous walks, GraphMixer~\cite{cong2023graphmixer} shows that simple temporal link encoders can be competitive, and DyGFormer~\cite{yu2023dygformer} uses historical first-hop interactions with Transformer-style sequence modeling. Frequency-oriented dynamic graph methods such as FreeDyG~\cite{tian2024freedyg} and SGD-DYG~\cite{han2025sgddyg} further explore global or spectral dependency modeling. Compared with these works, CacheDyG focuses on fixed-node snapshot sequences and treats reusable historical graph computation as the main efficiency opportunity. Its goal is not to introduce a heavier temporal encoder, but to cache temporal graph propagation and train only a lightweight correction and scoring module.

\subsection{Dynamic Graph Neural Networks}
Dynamic graph neural networks combine graph neural aggregation with temporal modeling to learn representations over evolving graph structures~\cite{wang2026gtcn}. Snapshot-based dynamic GNNs apply graph-based propagation across a sequence of graph snapshots. DySAT~\cite{sankar2020dysat} uses structural and temporal self-attention, EvolveGCN~\cite{pareja2020evolvegcn} evolves graph convolution parameters with recurrent networks, VGRNN~\cite{hajiramezanali2019vgrnn} introduces recurrent latent variables, and ROLAND~\cite{you2022roland} updates hierarchical node states over time. Other compact or convolutional dynamic GNN baselines, such as WinGNN~\cite{zhu2023wingnn} and GTCN~\cite{wang2024tensor}, also aim to model temporal graph evolution with different efficiency-accuracy trade-offs.

Continuous-time temporal GNNs represent dynamic graphs as timestamped interaction streams. TGAT~\cite{xu2020tgat} uses temporal attention for inductive representation learning, while TGN~\cite{rossi2020tgn} introduces memory modules and graph-based operators for event-driven dynamic graphs. These models are powerful for asynchronous temporal reasoning, but their design objectives differ from the transductive snapshot setting studied here. In snapshot-based training, the same historical graph structures are repeatedly reused across epochs, making cached propagation especially beneficial.

Unlike existing dynamic GNNs that couple temporal propagation with epoch-wise optimization, CacheDyG caches reusable graph computation and trains only lightweight refinement and prediction modules.

\subsection{Related Structure-Aware and Efficient Representation Learning}
Beyond canonical dynamic-graph architectures, recent work has explored complementary ways to represent evolving or spatiotemporal data. Dynamic graph mixers and spatiotemporal graph--tensor models explicitly couple temporal variation with relational structure~\cite{bi2025dynamicgraphmixer,bi2025stgnn,liao2025crypto,xu2026sampling,liao2025tensorcausal,wang2025dynamicdirected,xu2025dynamicqos,hou2025selfattending,wu2025spatiotemporal,yang2025traffic,chen2024dynamicqos}. These studies motivate the broader view that temporal dependency can be captured through graph, tensor, and latent-factor formulations, whereas CacheDyG specifically targets the repeated propagation cost that arises when a fixed sequence of graph snapshots is optimized over many epochs.

A second line of work develops structure-aware graph representation modules for different learning settings. Representative examples include modularized graph convolution~\cite{he2025modulargcn}, neural community search~\cite{lin2026ncsac}, attributed graph clustering~\cite{yang2025graphclustering}, graph-based multi-source association prediction~\cite{wu2025mirnadrug}, collaborative-distillation GNNs for recommendation~\cite{gou2026distillation}, graph linear convolution pooling~\cite{bi2025graphpool}, and graph-bi-regularized matrix factorization for community detection~\cite{liu2024graphnmf}. These methods emphasize how structural priors or task-specific graph operators can improve representation quality. CacheDyG is orthogonal to these designs: its main objective is to separate reusable graph propagation from routine parameter updates.

Efficiency has also been studied from compression, factorization, and optimization perspectives. Examples include low-rank tensor compression~\cite{he2026compression}, auto-encoding and neural Tucker factorization~\cite{tang2025autoencoding,tang2025neuraltucker}, surveys of parallel optimization for high-dimensional factorization~\cite{hu2025parallelreview}, attention-based and accelerated neural tensor factorization~\cite{xu2025attention,li2026accelerated}, fine-grained regularization of tensor factorization~\cite{wu2024regularization}, tensor-decomposition-based efficient detection~\cite{zeng2024efficientdetector}, adaptively accelerated parallel stochastic-gradient optimization~\cite{qin2024parallelsgd}, and controller- or population-based refinement of latent-factor learning~\cite{li2025pidrefine,yuan2024fuzzypid,lyu2026pso}. These works address efficiency or representation compactness through optimization and model design. In contrast, CacheDyG reduces redundant computation by amortizing sparse temporal graph propagation through a reusable non-trainable cache.

\section{Methodology}
\begin{figure}[t]
\centering
\includegraphics[width=0.96\linewidth]{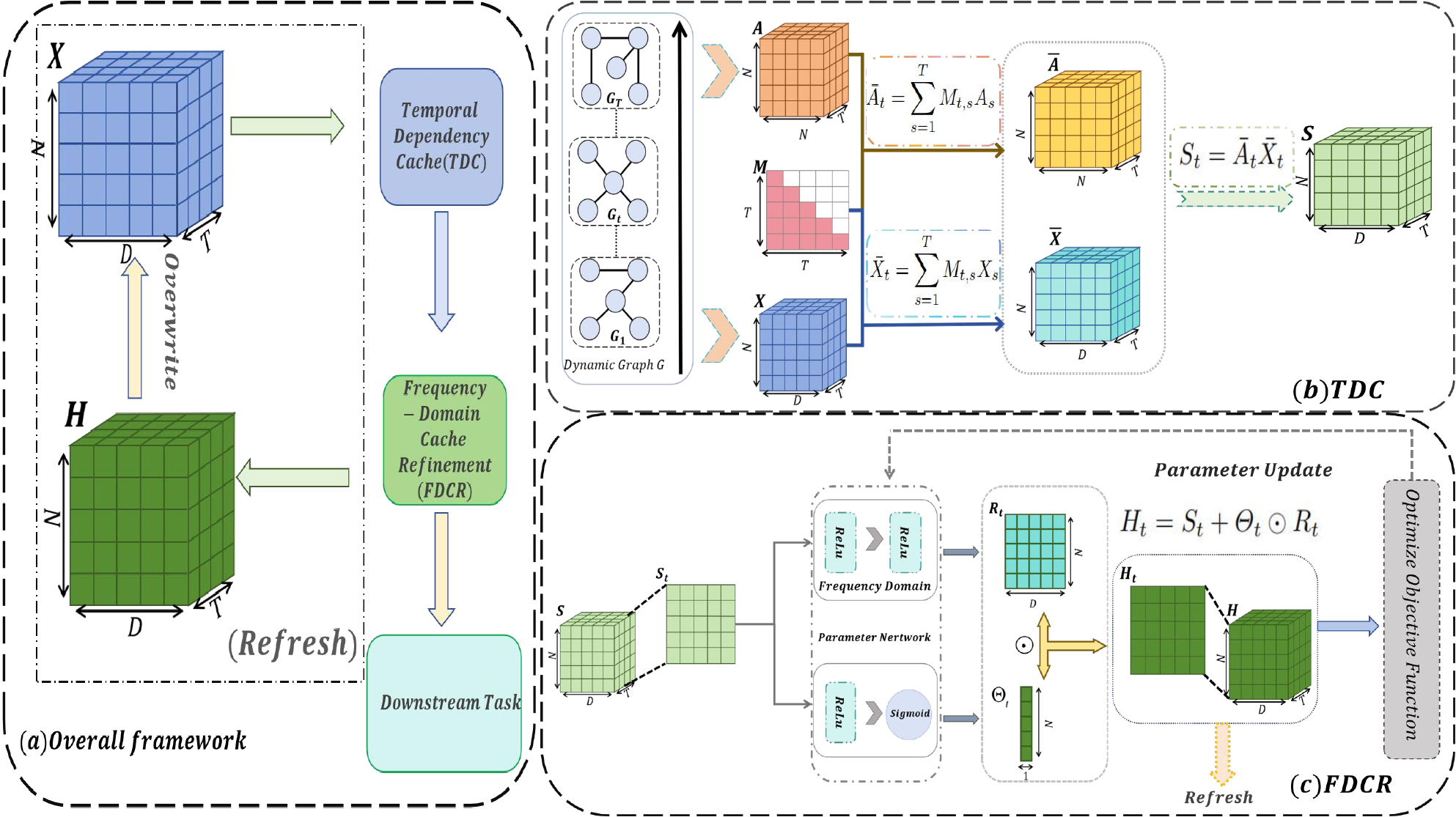}
\caption{Overall framework of CacheDyG. TDC constructs a temporally mixed graph-aware cache $S$; FDCR applies node-domain frequency refinement and residual gating; the refined representations are then used for link prediction while sparse propagation remains outside ordinary training epochs.}
\label{fig:framework}
\end{figure}

This section presents CacheDyG, whose design is summarized in Fig.~\ref{fig:framework}. The framework first builds a temporally mixed graph-aware cache, then adapts the cached representations with a lightweight node-domain frequency refiner for link prediction. In cache mode, sparse graph propagation is executed only during cache construction or refresh; ordinary epochs optimize only the refinement and prediction modules.

\subsection{Problem Formulation}
Let
\begin{equation}
\mathcal{G}=\{G_t\}_{t=1}^{T},\qquad G_t=(V,E_t),
\end{equation}
denote a sequence of graph snapshots, where $V=\{v_1,\ldots,v_N\}$ is a fixed node set after preprocessing and $E_t$ is the edge set observed at time step $t$. We use $B_t\in\mathbb{R}^{N\times N}$ to denote the raw adjacency matrix of $G_t$, and $A_t\in\mathbb{R}^{N\times N}$ to denote its normalized sparse adjacency matrix. A standard normalization with self-loops can be written as
\begin{equation}
A_t=D_t^{-\frac12}(B_t+I)D_t^{-\frac12},
\end{equation}
where $D_t$ is the degree matrix of $B_t+I$. For large sparse temporal graphs, the same formulation can be implemented with sparse-safe normalization without materializing dense self-loop structures.

The task is one-step link prediction. Given historical snapshots up to time $t$, the model predicts whether a candidate node pair $(u,v)$ will be connected in the next snapshot. The prediction target is
\begin{equation}
y_{t,u,v}=\begin{cases}
1, & (u,v)\in E_{t+1},\\
0, & (u,v)\notin E_{t+1},
\end{cases}
\end{equation}
and the model estimates
\begin{equation}
\hat y_{t,u,v}=p\left(y_{t,u,v}=1\mid \mathcal{G}_{\le t}\right).
\end{equation}
Positive samples are observed temporal edges, and negative samples are drawn from unobserved node pairs under the same temporal evaluation protocol.

\subsection{Temporal Dependency Cache}
CacheDyG represents temporal dependency using a temporal mixing matrix $M\in\mathbb{R}^{T\times T}$. The matrix is lower triangular, row stochastic, and band-limited:
\begin{equation}
M_{t,s}=0\quad\text{if }s>t,
\end{equation}
and each row aggregates information from the current snapshot and recent historical snapshots. More generally, the entries of $M$ are obtained by normalizing a non-negative decay kernel,
\begin{equation}
M_{t,s}=\frac{\kappa(t-s)\mathbf{1}[0\le t-s<b]}{\sum_{q=1}^{T}\kappa(t-q)\mathbf{1}[0\le t-q<b]},
\end{equation}
where $b$ controls the temporal support and $\kappa(\cdot)$ determines the relative importance of historical snapshots. This construction prevents future information leakage.

Let $X\in\mathbb{R}^{T\times N\times d}$ be the node cache tensor, where $X_t\in\mathbb{R}^{N\times d}$ stores the cached node representations at time step $t$. In cache mode, $X$ is maintained as a non-trainable buffer rather than as a directly optimized embedding table.

CacheDyG first constructs temporally mixed features and temporally mixed graph structures:
\begin{align}
\bar X_t &= \sum_{s=1}^{T} M_{t,s}X_s,\\
\bar A_t &= \sum_{s=1}^{T} M_{t,s}A_s.
\end{align}
The graph-aware cache is then obtained through sparse propagation:
\begin{equation}
S_t=\bar A_t\bar X_t,\qquad t=1,\ldots,T.
\end{equation}
The resulting tensor $S\in\mathbb{R}^{T\times N\times d}$ contains structural and temporal information. Since $\bar A_t$ depends only on the snapshot sequence and temporal mixing matrix, it can be precomputed and reused, moving the sparse propagation outside the inner gradient-update loop.

\subsection{Node-Domain Frequency-Domain Cache Refinement}
The cache $S$ encodes temporal propagation, but it is not directly optimized for the link prediction objective. CacheDyG therefore applies a lightweight refiner that calibrates the cached graph signal without performing additional sparse graph convolution in each epoch. The Fourier transform is applied along the node axis of each cache slice, following the node-axis spectral refinement convention used in lightweight frequency-based dynamic graph learning such as SGD-DYG~\cite{han2025sgddyg}. In CacheDyG, this operation is used as representation calibration after TDC rather than as a replacement for temporal modeling. For each time step $t$, the cache matrix $S_t$ is transformed along the node axis by a real-valued fast Fourier transform:
\begin{equation}
\widehat S_t=\mathcal{F}_{N}(S_t),
\end{equation}
where $\mathcal{F}_{N}(\cdot)$ denotes the Fourier transform over the node dimension and $\widehat S_t$ is complex-valued. The snapshot/time axis is kept fixed in this operation; temporal dependency has already been encoded by the temporal mixing and graph-aware propagation above. The transformed representation is processed by a compact complex-valued feed-forward network:
\begin{equation}
\widehat R_t=\Phi_{\Omega}(\widehat S_t),
\end{equation}
where $\Phi_{\Omega}$ consists of complex affine transformations and element-wise nonlinearities applied separately to the real and imaginary parts. The refined signal is mapped back to the original representation domain by the inverse transform:
\begin{equation}
R_t=\Psi\left(\mathcal{F}_{N}^{-1}(\widehat R_t)\right),
\end{equation}
where $\Psi(\cdot)$ is a real-valued projection. The resulting $R_t$ is therefore a node-domain correction term over a temporally mixed graph cache, not a separate temporal-frequency representation.

To preserve stable graph information, CacheDyG uses an adaptive residual gate computed from the cached representation:
\begin{equation}
\Theta_t=\sigma\left(g_{\Omega}(S_t)\right),
\end{equation}
where $g_{\Omega}$ is a small gating network and $\Theta_t\in(0,1)^{N\times 1}$ is broadcast over the feature dimension. The final refined representation is
\begin{equation}
H_t=S_t+\Theta_t\odot R_t,
\end{equation}
where $\odot$ denotes element-wise multiplication. The residual form keeps the cached graph signal as the default representation and injects node-domain corrections only when they improve the supervised objective.

\subsection{Link Prediction Objective}
For a candidate edge $(u,v)$ at time step $t$, CacheDyG extracts the corresponding refined node representations $h_{t,u}$ and $h_{t,v}$ from $H_t$. The link predictor is a lightweight binary classifier over the concatenated pair representation:
\begin{equation}
\hat y_{t,u,v}=\sigma\left(w^\top[h_{t,u}\Vert h_{t,v}]+b\right),
\end{equation}
where $\Vert$ denotes concatenation and $\sigma(\cdot)$ is the sigmoid function.

Given a training set $\mathcal{S}$ containing both positive and negative temporal node pairs, the model is optimized with the binary cross-entropy loss:
\begin{equation}
\mathcal{L}=-\frac{1}{|\mathcal{S}|}\sum_{(t,u,v)\in\mathcal{S}}\left[y_{t,u,v}\log \hat y_{t,u,v}+(1-y_{t,u,v})\log(1-\hat y_{t,u,v})\right].
\end{equation}
In the standard cache-mode loop, gradient descent updates only the refiner and predictor. The cached tensor is detached from the computational graph and is updated only through the refresh mechanism below.

\subsection{Cache-Refresh Training}
A completely fixed cache may become misaligned with the supervised objective. CacheDyG therefore refreshes the cache periodically or when validation performance stagnates. Let $\mathcal{R}_{\Omega}(\cdot)$ denote the refinement mapping above. When a refresh is triggered, the model first computes a detached refined cache:
\begin{equation}
Z=\operatorname{stopgrad}\left(\mathcal{R}_{\Omega}(S)\right),
\end{equation}
and then updates the non-trainable node cache by combining the previous cache with the refined representation:
\begin{equation}
X\leftarrow Z.
\end{equation}
After updating $X$, the graph-aware cache $S$ is rebuilt. Overall, cache-mode training builds $\{\bar A_t\}_{t=1}^{T}$, constructs $S$, optimizes the refiner and predictor with the binary cross-entropy objective, and refreshes $X$ and $S$ only when the refresh condition is met. The cache can therefore adapt during training while sparse propagation remains outside the inner loop.

\subsection{Computational Analysis}
Let $E=\sum_{t=1}^{T}|E_t|$ denote the total number of temporal edges and let $Q$ be the number of optimization epochs. A conventional dynamic graph neural network that performs sparse propagation over all snapshots in every epoch incurs a propagation cost proportional to
\begin{equation}
\mathcal{O}(QEd),
\end{equation}
up to model-dependent constants. By contrast, CacheDyG performs sparse propagation only when the cache is built or refreshed. If the cache is refreshed $R$ times during training, the sparse propagation cost becomes
\begin{equation}
\mathcal{O}(REd)+\mathcal{O}(Q C_{\mathrm{refine}}),
\end{equation}
where $C_{\mathrm{refine}}$ is the cost of the lightweight node-domain frequency refinement and link prediction modules. Since cache refresh is substantially less frequent than epoch-level optimization, the dominant sparse graph computation is amortized across training.

\section{Experiments}
We evaluate CacheDyG from four perspectives: link prediction accuracy, component contribution, refresh-schedule robustness, and the accuracy-efficiency trade-off against representative dynamic graph models.

\subsection{Dataset and Metrics}
We evaluate CacheDyG on five dynamic graph datasets: Wiki-Eo, Digg, Alpha, DBLP, and StackOverflow, covering Wikipedia interactions~\cite{han2025sgddyg}, social-news replies~\cite{kunegis2013konect}, signed trust relations~\cite{kumar2016edge}, academic relations~\cite{han2025sgddyg}, and Stack Exchange interactions~\cite{han2025sgddyg}. The datasets are discretized into 60, 50, 60, 45, and 25 snapshots, respectively. After preprocessing, all datasets use a fixed node set and chronological ordering. We adopt a 70\%/10\%/20\% train/validation/test split and predict links in snapshot $(t+1)$ given historical snapshots up to $t$.

For evaluation, each positive edge is paired with one unobserved node pair as a negative sample. We report average precision (AP) as the primary metric and ROC-AUC as a complementary ranking metric. MAP denotes the mean average precision over five random seeds, and MAUC denotes the mean ROC-AUC. Results are reported as mean and standard deviation over repeated runs. Trainable parameter counts exclude non-trainable cache buffers, and runtime is reported as training, validation, and total time, and a method that cannot complete under the same environment is marked as out-of-memory (OOM).

\subsection{Experiment Settings}
All methods use the same chronological split, negative samples, and evaluation protocol. CacheDyG is optimized with Adam using a learning rate of $1\times10^{-2}$, weight decay of $5\times10^{-4}$, a maximum of 200 epochs, and early stopping with a patience of 25. The random seeds are from 2024--2028. The input feature dimension is 8, the predictor contains one 16-dimensional hidden layer, and the temporal bandwidth is 20. The graph and frequency modules use a dropout rate of 0.75.

Unless otherwise specified, CacheDyG uses temporal mixing, a non-trainable $X_{\mathrm{cache}}$, cache graph mode, and the full node-domain frequency-domain refiner. The refiner and adaptive scaling networks have hidden dimensions 128 and 32, respectively. Thus, CacheDyG contains 19106 trainable parameters. Cache refresh is performed every 8 epochs for the first four warm-up refreshes and is subsequently triggered after 10 epochs without validation improvement. During ordinary epochs, only the refiner and link predictor are optimized; sparse graph propagation is executed only when the cache is constructed or refreshed.

\subsection{Baselines}
We compare CacheDyG with six representative baselines: DySAT for structural-temporal self-attention, ROLAND~\cite{you2022roland} for rolling node-state updates, EvolveGCN for recurrent evolution of graph convolution parameters, WinGNN~\cite{zhu2023wingnn} for compact temporal interaction modeling, GTCN~\cite{wang2024tensor} for graph-temporal convolution, and SGD-DYG~\cite{han2025sgddyg} for lightweight frequency-based dynamic graph learning.

\subsection{Comparative Performance Analysis}
Table~\ref{tab:main} reports the link prediction performance on all five datasets.

\begin{table*}[t]
\centering
\caption{Main link prediction results on dynamic graph datasets.}
\label{tab:main}
\resizebox{\textwidth}{!}{%
\begin{tabular}{llrrrrrrr}
\toprule
Dataset & Metric & DySAT & ROLAND & EvolveGCN & WinGNN & GTCN & SGD-DYG & CacheDyG\\
\midrule
Wiki-Eo & MAP & 85.70$\pm$0.11 & 83.90$\pm$0.96 & 83.88$\pm$3.47 & 82.16$\pm$4.15 & 92.64$\pm$1.03 & 92.71$\pm$1.24 & \textbf{95.14$\pm$1.28}\\
& MAUC & 81.26$\pm$0.71 & 82.52$\pm$1.37 & 74.47$\pm$6.20 & 81.21$\pm$4.07 & 91.12$\pm$1.64 & 91.32$\pm$1.52 & \textbf{94.82$\pm$0.57}\\
Digg & MAP & 77.10$\pm$0.38 & 73.30$\pm$1.61 & 74.56$\pm$2.31 & 72.75$\pm$1.69 & 75.86$\pm$0.77 & 76.76$\pm$0.71 & \textbf{77.87$\pm$0.30}\\
& MAUC & 74.10$\pm$0.50 & 71.78$\pm$1.16 & 69.86$\pm$3.26 & 68.89$\pm$1.71 & 73.25$\pm$0.66 & 74.49$\pm$1.84 & \textbf{74.51$\pm$0.94}\\
Alpha & MAP & 86.89$\pm$2.22 & 86.98$\pm$7.19 & 86.60$\pm$1.98 & 87.38$\pm$3.46 & 91.71$\pm$0.56 & 91.86$\pm$0.09 & \textbf{93.02$\pm$1.08}\\
& MAUC & 85.39$\pm$2.36 & 83.96$\pm$7.90 & 79.53$\pm$2.02 & 81.54$\pm$2.23 & 92.39$\pm$0.60 & \textbf{93.89$\pm$0.37} & 93.17$\pm$0.98\\
DBLP & MAP & OOM & 61.81$\pm$0.56 & 63.46$\pm$3.73 & 62.32$\pm$0.67 & 61.37$\pm$0.58 & 60.87$\pm$0.27 & \textbf{67.32$\pm$0.59}\\
& MAUC & OOM & 61.78$\pm$0.63 & 53.64$\pm$4.70 & \textbf{62.92$\pm$0.90} & 62.47$\pm$0.63 & 60.95$\pm$0.33 & 57.38$\pm$4.81\\
StackOverflow & MAP & OOM & 88.01$\pm$1.33 & 89.07$\pm$2.16 & 74.34$\pm$1.08 & 90.28$\pm$0.88 & 90.38$\pm$0.84 & \textbf{92.05$\pm$0.92}\\
& MAUC & OOM & 85.21$\pm$2.59 & 86.30$\pm$1.84 & 73.44$\pm$2.83 & 88.44$\pm$0.93 & 88.46$\pm$1.27 & \textbf{89.67$\pm$1.03}\\
\bottomrule
\end{tabular}}
\end{table*}

\paragraph{Overall effectiveness.} Table~\ref{tab:main} shows that CacheDyG achieves the best AP on all five datasets, with gains of 2.43, 0.77, 1.16, 3.86, and 1.67 percentage points over the strongest AP baseline on Wiki-Eo, Digg, Alpha, DBLP, and StackOverflow, respectively. The ROC-AUC results are also competitive: CacheDyG ranks first on Wiki-Eo, Digg, and StackOverflow and second on Alpha. On DBLP, the AP gain is clear while ROC-AUC is lower than the best baseline, suggesting that the method is especially strong for positive-edge ranking but not uniformly dominant under every ranking metric.

\paragraph{Scalability evidence.} The results on DBLP and StackOverflow are particularly relevant to efficiency. DySAT runs out of memory on both datasets, whereas CacheDyG remains feasible and achieves the highest AP. This supports the main design goal: cached temporal propagation and lightweight refinement can preserve predictive quality when repeated sparse propagation becomes costly. We evaluate efficiency through trainable parameter size and wall-clock runtime. Table~\ref{tab:param} reports trainable parameters in thousands, and Tables~\ref{tab:runtime1}--\ref{tab:runtime2} report training, validation, and total runtime. StackOverflow is listed separately because of its larger scale.

\begin{table}[t]
\centering
\caption{Trainable parameter comparison on all datasets. Values are in K.}
\label{tab:param}
\resizebox{\linewidth}{!}{\begin{tabular}{lrrrrr}
\toprule
Model & Wiki-Eo & Digg & Alpha & DBLP & StackOverflow\\
\midrule
DySAT & 2462.656 & 8328.944 & 2566.304 & OOM & OOM\\
ROLAND & 69.409 & 245.865 & 63.825 & 2884.009 & 20676.305\\
EvolveGCN & 2393.977 & 8261.161 & 2497.625 & 89315.009 & 702896.121\\
WinGNN & 59.408 & 243.976 & 61.936 & 2882.120 & 20673.872\\
GTCN & 2390.689 & 8257.937 & 2494.337 & 89311.721 & 381446.465\\
SGD-DYG & 2390.753 & 8257.873 & 2494.401 & 89311.785 & 351446.465\\
CacheDyG & \textbf{19.106} & \textbf{19.106} & \textbf{19.106} & \textbf{19.106} & \textbf{19.106}\\
\bottomrule
\end{tabular}}
\end{table}

\begin{table}[t]
\centering
\caption{Train, validation, and total runtime of one epoch on non-StackOverflow datasets. Values are in seconds; each cell reports train/validation/total.}
\label{tab:runtime1}
\resizebox{\linewidth}{!}{\begin{tabular}{lrrrr}
\toprule
Model & Wiki-Eo & Digg & Alpha & DBLP\\
\midrule
DySAT & 0.142/0.067/0.210 & 0.287/0.083/0.370 & 0.157/0.045/0.202 & OOM\\
ROLAND & 0.234/0.132/0.367 & 0.195/0.103/0.299 & 0.235/0.120/0.355 & 0.259/0.137/0.396\\
EvolveGCN & 0.288/0.063/0.351 & 0.378/0.080/0.458 & 0.299/0.046/0.345 & 0.465/0.065/0.530\\
WinGNN & 0.204/0.057/0.261 & 0.217/0.050/0.267 & 0.132/0.039/0.171 & 0.308/0.097/0.405\\
GTCN & 0.105/0.030/0.135 & 0.133/0.062/0.195 & 0.111/0.023/0.134 & 0.302/0.102/0.434\\
SGD-DYG & 0.197/0.114/0.312 & 0.205/0.133/0.338 & 0.202/0.117/0.319 & 1.330/0.452/1.780\\
CacheDyG & \textbf{0.005/0.002/0.007} & \textbf{0.009/0.005/0.014} & \textbf{0.006/0.002/0.008} & \textbf{0.075/0.050/0.125}\\
\bottomrule
\end{tabular}}
\end{table}

\begin{table}[t]
\centering
\caption{Train, validation, and total runtime of one epoch on StackOverflow. Values are in seconds.}
\label{tab:runtime2}
\begin{tabular}{lrrr}
\toprule
Model & Train time & Validation time & Total time\\
\midrule
DySAT & OOM & OOM & OOM\\
ROLAND & 92.058 & 38.221 & 130.280\\
EvolveGCN & 164.448 & 24.947 & 189.395\\
WinGNN & 84.974 & 30.513 & 115.486\\
GTCN & 88.541 & 75.793 & 164.334\\
SGD-DYG & 304.999 & 180.505 & 485.503\\
CacheDyG & \textbf{65.902} & \textbf{31.402} & \textbf{97.304}\\
\bottomrule
\end{tabular}
\end{table}

\paragraph{Parameter efficiency.} Table~\ref{tab:param} shows that CacheDyG uses 19.106K trainable parameters on every dataset because node-time representations are stored as non-trainable buffers. On StackOverflow, the smallest baseline already exceeds 20,000K parameters, more than 1,082$\times$ the size of CacheDyG.

\paragraph{Runtime efficiency.} On Wiki-Eo, Digg, Alpha, and DBLP, CacheDyG obtains the lowest total runtime, with speedups of approximately 19.3$\times$, 13.9$\times$, 16.8$\times$, and 3.2$\times$ over the fastest baseline on each dataset. On StackOverflow, it completes in 97.304 seconds, lower than WinGNN's 115.486 seconds, while also achieving the highest AP. Together, the accuracy and efficiency results support cache-refine learning for transductive snapshot sequences in which historical graph structures are repeatedly reused.

\paragraph{Effectiveness of the Lightweight Design.} CacheDyG removes redundant trainable capacity without discarding useful graph information. The TDC stores reusable graph representations, the non-trainable cache avoids node-time overparameterization, and the FDCR provides task-specific correction with few parameters. Residual gating and selective refresh preserve stable cached signals while maintaining alignment with the prediction objective, explaining the favorable accuracy-efficiency trade-off.

\subsection{Ablation Study}
We conduct ablation experiments on Wiki-Eo to isolate the contribution of each design choice. The variants remove cache-style propagation, remove the refiner, make $X_{\mathrm{cache}}$ trainable, remove temporal mixing, remove cache refresh, or keep graph convolution inside the refinement stage (TGC Graph Mode). All variants use the same data split and evaluation protocol.

\begin{table}[t]
\centering
\caption{Ablation study on Wiki-Eo. Parameter values are in K, with multiples relative to the full CacheDyG in parentheses.}
\label{tab:ablation}
\resizebox{\linewidth}{!}{\begin{tabular}{lrrrr}
\toprule
Model variant & AP & ROC-AUC & Parameters & Train time (s)\\
\midrule
Full CacheDyG & 96.14 & 96.11 & 19.106 (1.0$\times$) & $7.0\times10^{-3}$\\
w/o Cache-style Prop. & 87.46 & 84.42 & 2390.753 (125.1$\times$) & $4.8\times10^{-2}$\\
w/o Refiner & 50.80 & 57.47 & 0.017 (0.001$\times$) & $2.0\times10^{-3}$\\
Trainable $X_{\mathrm{cache}}$ & 92.69 & 91.21 & 2401.226 (125.7$\times$) & $5.2\times10^{-2}$\\
w/o Temporal Mixing & 90.31 & 86.37 & 5.434 (0.28$\times$) & $5.0\times10^{-3}$\\
w/o Cache Refresh & 81.52 & 78.73 & 19.106 (1.0$\times$) & $7.0\times10^{-3}$\\
TGC Graph Mode & 94.39 & 92.13 & 22.642 (1.2$\times$) & $5.3\times10^{-2}$\\
\bottomrule
\end{tabular}}
\end{table}

\paragraph{Component effects.} Table~\ref{tab:ablation} shows that removing cache-style propagation substantially reduces AP and ROC-AUC while increasing the parameter count by more than two orders of magnitude. Removing the refiner causes the largest performance drop, indicating that the raw cache must be calibrated toward the link prediction objective. Making $X_{\mathrm{cache}}$ trainable also underperforms the full model despite using far more parameters, which supports the use of a non-trainable cache with compact refinement.

\paragraph{Temporal dependency and decoupling.} Removing temporal mixing or cache refresh reduces performance, confirming that both cross-snapshot dependency and adaptive cache evolution are important. TGC Graph Mode remains competitive but is slower and uses more parameters than the full cache graph mode. Thus, graph-aware propagation is useful, but it need not be executed inside every ordinary training epoch.

\subsection{Parameter Sensitivity Analysis}
We visualize the sensitivity of the cache-refresh schedule on Wiki-Eo. The experiment varies warm-up refreshes, refresh interval, and endurance threshold over $\{2,4,6,8,10\}$, yielding 125 combinations with all other settings fixed. This analysis is dataset-specific and is not intended to claim that the same landscape holds across all datasets.

\begin{figure}[t]
\centering
\includegraphics[width=0.95\linewidth]{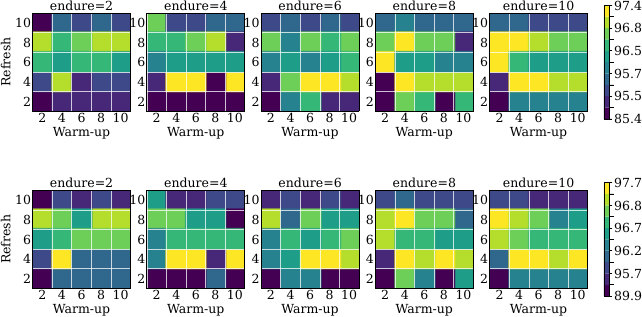}
\caption{Refresh-schedule heatmaps on Wiki-Eo. The top row reports MAUC and the bottom row reports MAP.}
\label{fig:refresh}
\end{figure}

\begin{figure}[t]
\centering
\includegraphics[width=0.72\linewidth]{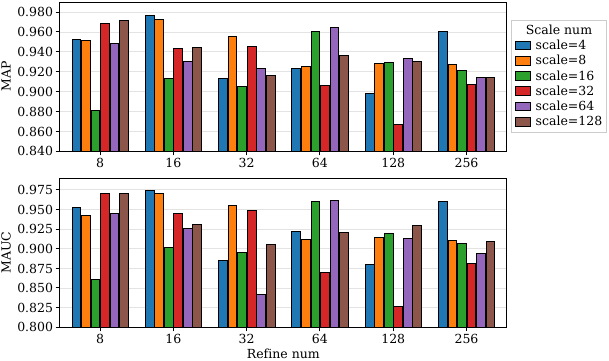}
\caption{Sensitivity of MAP and MAUC to the refinement number and scale number.}
\label{fig:sensitivity}
\end{figure}

Fig.~\ref{fig:refresh} shows a broad high-performance region rather than a single isolated optimum. Most moderate configurations maintain AP and ROC-AUC around or above 0.96, while extreme refresh schedules are more likely to reduce performance. Fig.~\ref{fig:sensitivity} further examines the interaction between the refinement number and scale number. The MAP and MAUC bars remain high across multiple moderate combinations, indicating that the cache-refinement module is not dependent on a single fragile hyperparameter choice. Very large or imbalanced settings can introduce unnecessary correction noise, but the overall trend supports a stable middle range for refinement and scaling. On Wiki-Eo, cache refresh and cache refinement are therefore not highly sensitive once the hyperparameters remain in a moderate range.

\section{Conclusion}
This paper presented CacheDyG, which decouples temporal graph propagation from routine parameter updates by storing graph-aware node-time representations in non-trainable caches and optimizing only lightweight refinement and prediction modules. Selective refresh keeps the cache aligned with the supervised objective. Experiments on five datasets demonstrate competitive link-prediction accuracy, substantially fewer trainable parameters, and lower runtime, supporting cache-based decoupling as an efficient design for fixed-node snapshot dynamic graphs.

\end{document}